\documentclass[11pt]{article}

\usepackage[T1]{fontenc}
\usepackage{lmodern}
\usepackage[margin=1in]{geometry}
\usepackage{amsmath,amssymb}
\usepackage{booktabs}
\usepackage{graphicx}
\usepackage{url}
\usepackage[hidelinks]{hyperref}

\title{Beyond Pass@k: Measuring Reliability and Security of Agentic Code Generation}

\author{%
Jiajun Jiang$^{1}$ \quad
Sharon Zheng$^{1}$ \quad
Natan Vidra$^{2}$ \quad
Spurthi Setty$^{3}$\\[0.6em]
\small $^{1}$Cornell University \quad
$^{2}$Anote.Ai \quad
$^{3}$Stevens Institute of Technology
}
\date{}

\begin{document}

\maketitle

\begin{abstract}
AI coding agent benchmarks rank agents with the Chen et al. (2021) pass@k estimator, but current implementations misapply it: they set $n$ to the number of unit tests in a single submission rather than the number of independent rollout attempts, conflating test-suite size with attempt independence. We diagnose this operationalization error, prove it by counterexample, and propose reliability@k---the same estimator applied correctly, with $n$ = independent rollouts and $c$ = fully-passing rollouts per (task, agent) pair. In a synthetic multi-rollout benchmark, the misapplied metric inflates reported scores by 0.85--0.97 in absolute terms (0.96--0.98 reported vs. 0.00--0.12 corrected), and a cheap single-rollout proxy fails to substitute for repeated runs (Spearman $\rho = 0.417$). Motivated by evidence that functional correctness does not imply security safety, we additionally propose security-adjusted reliability@k, which counts only rollouts that are both functionally correct and free of high-severity insecure patterns; in an initial live-API test with three agents, the adjustment did not change any ranking under our current scanner and threshold, so we present it as a proposed complementary lens whose decisive evaluation requires better-powered future runs. Finally, a preliminary 5-task SWE-bench Verified pilot observes the same core concern in a real repository setting: macro-averaged hidden-test pass rate was 0.80 while strict task resolution was 0.20.
\end{abstract}

\section{Introduction}
AI coding agents are increasingly deployed in production software engineering workflows, with tools such as Claude Code, GitHub Copilot, and Codex handling a multitude of coding tasks from implementation to bug fixes. The role of benchmarks to evaluate them and the metrics used to rank them become crucial in the decision of agent usage. A flawed benchmark does not only falsify performance, but also misdirects development priorities and misleads enterprises making deployment decisions. We will be studying the impact that reliability and security has on agentic AI deployment.

This paper identifies two independent failures in the way the field currently evaluates AI coding agents.

The first failure is mathematically-based. The evaluation metric, pass@k, is based on the unbiased combinatorial estimator introduced by Chen et al.~\cite{chen2021}, which estimator is mathematically correct, but requires $n$ independent, identically distributed full-solution samples as input. Current practice instead substitutes the number of unit tests in a single submission as $n$, and the number of passing tests as $c$. We found these are not independent samples; they are correlated sub-results of a single execution. Consequently, pass@k scores depend on the test-suite size rather than agent capability. Two agents with identical 40 percent task success rates receive scores of 0.976 and 1.000 respectively, purely because their test suites have different numbers of tests. Across a benchmark, this inflates reported scores by 0.85--0.97 in absolute terms and, in principle, can invert agent rankings.

The second failure is a security blind spot. Even when functional correctness is measured accurately, it captures only whether code passes tests---not whether it is safe to deploy. Veracode (2025)~\cite{veracode2025} reports that AI-generated code introduces $2.74\times$ more vulnerabilities than human-written code, and approximately 45 percent of AI-generated solutions that pass all functional tests contain high-severity security flaws such as SQL injection (CWE-89), OS command injection (CWE-78), or unsafe deserialization (CWE-502).

We present CodeBench, an evaluation framework that addresses both failures. We make the following contributions:
\begin{itemize}
  \item \textbf{Formal proof of the pass@k operationalization error.} We demonstrate that substituting unit-test counts for independent rollout counts violates the i.i.d. assumption of the Chen et al. estimator, and provide a counterexample showing the resulting score differences on agents with identical true reliability (H1).
  \item \textbf{reliability@k.} We propose the correct operationalization of the Chen et al. formula: applying it with $n$ = independent rollouts per (task, agent) and $c$ = rollouts achieving full execution success. In a synthetic multi-rollout benchmark, this reduces reported scores from 0.96--0.98 to 0.00--0.12, radically altering absolute score interpretation even where ranking order is preserved (H2).
  \item \textbf{Empirical validation that single-rollout proxies are insufficient.} We test whether a cheap proxy---pass rate weighted by regression penalty and tool efficiency---correlates strongly enough with reliability@k to substitute for repeated evaluation. It does not (Spearman $\rho < 0.70$), establishing that multi-rollout data collection is necessary for trustworthy reliability estimates (H3).
  \item \textbf{security\_adjusted\_reliability@k.} We propose a compound metric that conditions reliability@k on heuristic security screening, motivated by the concern that an enterprise selecting the most reliable agent may simultaneously be selecting the most vulnerable code generator. In an initial live-API test with three agents, the adjusted metric left every ranking unchanged under our current scanner and lenient threshold (Kendall $\tau = 1.000$); we therefore present it as a proposed complementary lens whose decisive test requires a better-powered evaluation (H4).
\end{itemize}

Beyond these contributions, we report a preliminary external validation: a 5-task SWE-bench Verified pilot (Section~\ref{sec:swepilot}) in which strict task-level resolution (0.20) diverged sharply from the macro-averaged hidden-test pass rate (0.80), consistent with the estimator concerns above in a real repository setting.

\section{Background}
\subsection{LLM Code Generation and Sampling}
Large language models generate code by predicting a probability distribution over the next token given a prompt and all previously generated tokens. At each decoding step, the model assigns a probability to every token in its vocabulary and samples from that distribution to produce an output. The key parameter governing this process is temperature: at temperature 0 (greedy decoding), the model always selects the highest-probability token, producing deterministic output. As temperature increases, the distribution flattens, making lower-probability tokens more likely to be sampled and introducing stochasticity into the output. This means the same prompt submitted twice at nonzero temperature will produce different completions---sometimes subtly (a variable renamed, a loop restructured) and sometimes drastically (an entirely different algorithmic approach). This run-to-run variance is not a bug; it reflects genuine uncertainty in the model's output distribution and is the source of the variance that pass@k is designed to measure. By sampling multiple completions and checking how many pass tests, pass@k estimates the probability that at least one of $k$ independent draws from the model's distribution produces a correct solution.

\subsection{The pass@k Metric}
Chen et al.~\cite{chen2021} introduced pass@k in the Codex paper as a way to evaluate LLM code generation without the high variance of a naive estimator. The naive approach---generate $k$ samples, check how many pass, divide---has high variance when $k$ is small. Instead, Chen et al. proposed an unbiased estimator: generate $n$ samples ($n \ge k$), count $c$ correct ones, and compute:
\begin{equation}
\mathrm{pass@}k = \mathbb{E}\left[1 - \frac{\binom{n-c}{k}}{\binom{n}{k}}\right].
\end{equation}
This estimates the probability that at least one of $k$ randomly chosen samples from the $n$ generated is correct. The formula rests on two critical assumptions: first, that each of the $n$ samples is an independent draw from the model's output distribution; second, that $n$ is reasonably large relative to $k$ to stabilize the estimate. These assumptions are satisfied when $n$ is the number of independent code completions sampled from the model for the same prompt. They are violated when $n$ is reinterpreted as the number of test cases in a single submission---a category error that conflates test-suite diversity with attempt independence and is the central problem this paper addresses.

\subsection{The Code Generation Benchmark Landscape}
Early code generation benchmarks such as HumanEval~\cite{chen2021} and MBPP~\cite{austin2021} evaluate models on short, self-contained programming tasks---typically single functions solvable in under 20 lines of code. These benchmarks are naturally suited to pass@k: generating 10--20 independent completions of a short function is cheap, fast, and the outputs are genuinely independent draws from the model's distribution. As the field matured, benchmarks grew in scope and complexity. SWE-bench~\cite{jimenez2024} evaluates agents on real GitHub issues requiring multi-file edits, dependency awareness, and test suite navigation across entire repositories. EvalPlus~\cite{liu2023} augments HumanEval with additional test cases to reduce false positives but inherits the same single-function evaluation paradigm. LiveCodeBench~\cite{jain2024} addresses training data contamination through time-stratified problems but does not change the evaluation protocol. CodeBench sits in the SWE-bench tier: tasks are multi-step, repo-level, and expensive to run. This distinction matters because the evaluation methodology appropriate for HumanEval does not transfer to this setting.

\subsection{Agentic vs. Completion-Based Code Generation}
Completion-based code generation treats the model as a stateless function: given a prompt, produce a completion. Each call is independent, cheap, and fast---ideal for sampling-based evaluation. Agentic code generation is fundamentally different. A coding agent operates in a loop: it reads the task, forms a plan, issues tool calls (file reads, terminal commands, test runners, web searches), observes the results, updates its state, and iterates until it judges the task complete or exhausts its budget. Each full agent run on a single task may involve dozens of tool calls, consume significant time and API tokens, and produce side effects in the environment that must be reset between runs. This makes each attempt expensive and deliberate rather than a cheap independent sample. An engineer deploying a coding agent in production does not run it fifty times on the same task and pick the best output---they run it once and need it to work. This operational reality is what reliability@k is designed to measure: not the probability that at least one of many cheap samples is correct, but the probability that a single deliberate attempt succeeds, estimated from a modest number of repeated runs.

\section{Experimental Setup}
\subsection{Benchmark and Task Corpus}
CodeBench's task corpus consists of ten curated coding problems spanning algorithmic, data-structure, and lightweight repository-style scenarios. The tasks span three tiers:
\begin{itemize}
  \item \textbf{Easy (3 tasks):} import parsing (AST-based), merge sort, and single-function implementations.
  \item \textbf{Medium (3 tasks):} data pipeline design, LRU cache, and Trie data structure.
  \item \textbf{Hard (4 tasks):} Dijkstra's algorithm, BM25 ranking, knapsack DP, and streaming median finder.
\end{itemize}
Each CodeTask has a natural-language description, reference solution in Python, and a corresponding test file path. For the synthetic benchmark experiments (seed=42), these tasks are paired with synthetic agent submissions that simulate repeated evaluation runs.

\subsection{Agents and Submissions}
We evaluate three synthetic agent profiles labeled:
\begin{itemize}
  \item \textbf{agent-x:} Agent profile 1.
  \item \textbf{claude-code:} Agent profile 2.
  \item \textbf{codex:} Agent profile 3.
\end{itemize}
\emph{Note: This is a synthetic benchmark experiment, not an evaluation of production systems.} Submissions are generated synthetically via \texttt{make\_rollout\_benchmark()} which creates independent rollouts per (task, agent) pair. Each rollout is modeled with a latent ``true skill'' parameter $s \in [0.30, 1.0]$, sampled uniformly, and per-rollout performance is drawn from a truncated Gaussian centered at $s$ with variance proportional to $(1.0 - 0.5s)$. This distribution is designed to introduce controlled rollout-level variance:
\begin{itemize}
  \item Rollout-level variation across repeated runs.
  \item A continuous latent skill parameter that determines long-run reliability.
  \item Deterministic reproducibility via seed control (seed=42).
\end{itemize}

\subsection{Evaluation Metrics}
CodeBench implements five core evaluation metrics for synthetic submissions.

\paragraph{Pass Rate.}
\begin{equation*}
\mathrm{pass\_rate} = \frac{\mathrm{tests\_passed}}{\mathrm{tests\_total}}.
\end{equation*}
Fraction of unit tests passing in a single submission.

\paragraph{Regression Rate.}
\begin{equation*}
\mathrm{regression\_rate} = \frac{\mathrm{regression\_count}}{\mathrm{tests\_total}}.
\end{equation*}
Fraction of test failures attributed to regressions (introducing bugs in working test cases).

\paragraph{Tool Efficiency Score.}
\begin{equation*}
\mathrm{tool\_efficiency} = \max\left(0, 1 - \frac{\mathrm{tool\_calls\_used}}{20}\right).
\end{equation*}
Penalizes excessive tool invocations; designed to reward concise agent behavior.

\paragraph{Cost-Adjusted Score.}
\begin{equation*}
\mathrm{cost\_adjusted} = \frac{\mathrm{pass\_rate}}{\log(1 + \mathrm{cost\_usd})}.
\end{equation*}
Normalizes pass rate by log cost, rewarding efficient solutions.

\paragraph{Security Score.}
A heuristic scanner detecting seven high-severity patterns in generated code: \texttt{eval()}, \texttt{exec()}, \texttt{os.system()}, \texttt{subprocess} with \texttt{shell=True}, \texttt{pickle.load()}, \texttt{yaml.load()} without \texttt{Loader}, and \texttt{\_\_import\_\_()}. Each pattern reduces the score by $1/7$; return value is in $[0,1]$.

\subsection{Experimental Protocol}
All experiments use a fixed seed (seed=42) for reproducibility. The primary experimental harness is a synthetic benchmark constructed via \texttt{make\_rollout\_benchmark(n\_tasks=10, agents=3, n\_rollouts=8, seed=42)}, which yields $10 \times 3 \times 8 = 240$ synthetic \texttt{ExecutionResult} entries. Each agent profile has a latent ``true skill'' parameter (sampled $\sim U[0.30, 1.0]$), and per-rollout outcomes are drawn from a truncated Gaussian around this skill value. For each (task, agent) pair, \texttt{execution\_success} is set to True if the task achieves near-complete test success (\texttt{pass\_rate} $\ge 0.95$) in that rollout.

Experiments 0--3 are run sequentially on this synthetic benchmark:
\begin{itemize}
  \item \textbf{Exp 0:} Baseline leaderboard using the broken \texttt{\_estimate\_pass@k} metric (conflating unit-test counts with rollout counts).
  \item \textbf{Exp 1:} Controlled proof of i.i.d. violation (H1). Two synthetic agent profiles with identical 0.40 single-submission test pass rates but different test-suite sizes receive different test-case-based pass@5 scores.
  \item \textbf{Exp 2:} Score inflation magnitude under correct vs. broken pass@k (H2). Measures gap between \texttt{\_estimate\_pass@k} and reliability@k on the synthetic benchmark.
  \item \textbf{Exp 3:} Correlation analysis of single-rollout proxy vs. reliability@5 (H3). Computes Spearman's $\rho$ on all 30 (task, agent) pairs from the synthetic data.
\end{itemize}

\paragraph{Real-world pilot setup.}
In addition to the synthetic benchmark (H1--H3) and the live-API H4 run, we conducted a small real-world pilot on SWE-bench Verified~\cite{jimenez2024}: the first 5 deterministic instances of the test split, one attempt per task by a single agent (Claude Code CLI v2.1.185, \texttt{acceptEdits} permission mode), with patches evaluated by the official SWE-bench harness in Docker. We frame this as a preliminary external validation rather than a full benchmark; results appear in Section~\ref{sec:swepilot}.

\section{Results}
\subsection{Pass@k Operationalization Error (H1)}
\begin{figure}[t]
  \centering
  \includegraphics[width=0.88\linewidth]{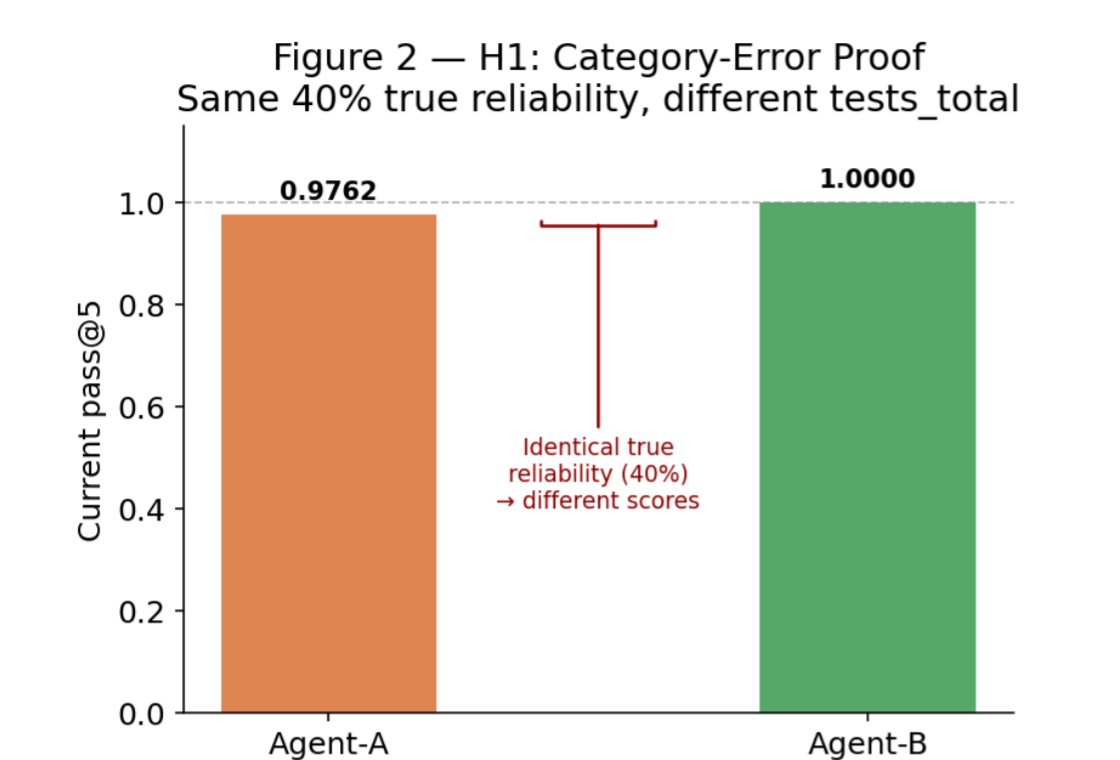}
  \caption{Category-error proof for the current pass@5 implementation. Two single submissions have the same unit-test pass rate but receive different pass@5 scores because the implementation treats unit tests as independent samples. Synthetic CodeBench experiment (seed=42).}
  \label{fig:category}
\end{figure}

Experiment 1 directly tests the core measurement error. We construct two synthetic agent profiles within CodeBench:
\begin{itemize}
  \item \textbf{Agent-A:} 1 submission with 10 unit tests; 4 pass. Single-submission test pass rate: $4/10 = 0.40$.
  \item \textbf{Agent-B:} 1 submission with 5 unit tests; 2 pass. Single-submission test pass rate: $2/5 = 0.40$.
\end{itemize}
Applying the broken \texttt{\_estimate\_pass@k()} formula (treating $n$ = test count, $c$ = tests passed):
\begin{align}
\mathrm{pass@5}_{\mathrm{broken}}(\mathrm{Agent\text{-}A})
&= 1 - \frac{\binom{10-4}{5}}{\binom{10}{5}} = 0.9762, \\
\mathrm{pass@5}_{\mathrm{broken}}(\mathrm{Agent\text{-}B})
&= 1 - \frac{\binom{5-2}{5}}{\binom{5}{5}} = 1.0000.
\end{align}
Despite identical 0.40 single-submission test pass rates, Agent-B scores 0.0238 points higher due to test-suite size alone. This violates the i.i.d. assumption of the Chen et al. (2021) estimator and confirms H1. Figure~\ref{fig:category} visualizes this category error.

\subsection{Reliability@k Results (H2)}
Experiment 2 measures the inflation of the test-case-based pass@5 metric (breaking the i.i.d. assumption) compared to the correct multi-rollout reliability@5. Using 8 independent rollouts per (task, agent) pair:

\begin{table}[t]
\centering
\caption{Test-case-based pass@5 substantially inflates corrected reliability@5 in the synthetic CodeBench benchmark.}
\label{tab:inflation}
\begin{tabular}{lrrr}
\toprule
Agent & Pass@5 & Rel.@5 & Infl. \\
\midrule
agent-x & 0.975 & 0.125 & 0.850 \\
claude-code & 0.965 & 0.000 & 0.965 \\
codex & 0.959 & 0.000 & 0.959 \\
\midrule
Mean & 0.966 & 0.042 & 0.924 \\
\bottomrule
\end{tabular}
\end{table}

\begin{figure}[t]
  \centering
  \includegraphics[width=0.88\linewidth]{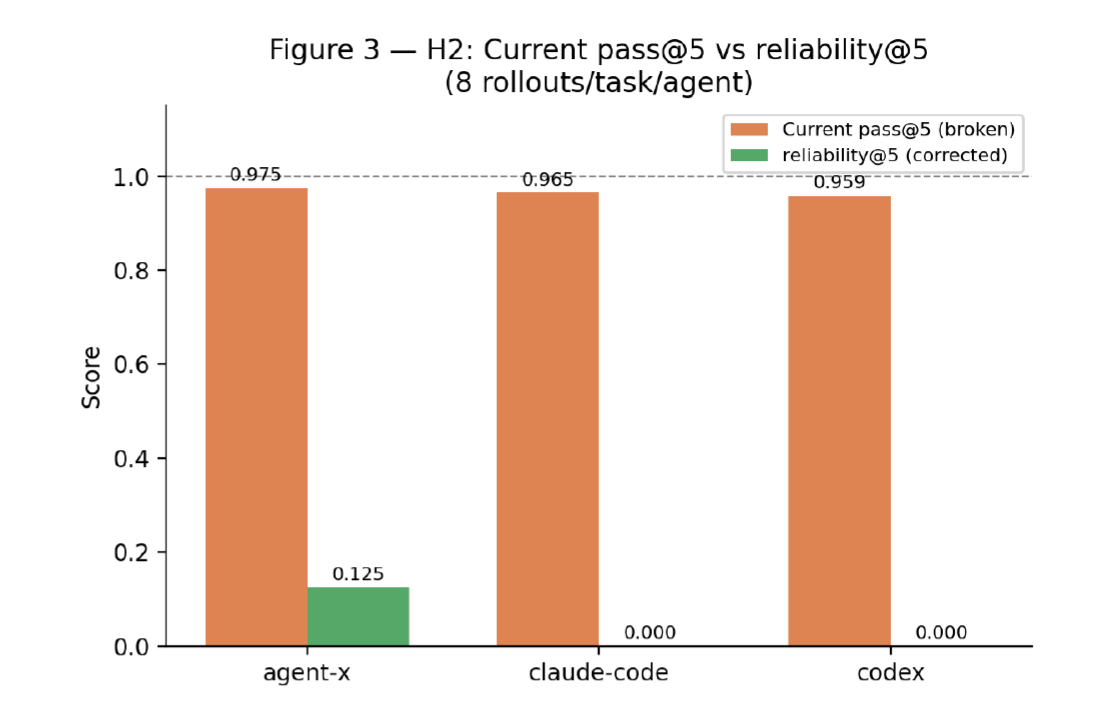}
  \caption{Score inflation under current vs. corrected pass@k implementation. Test-case-based pass@5 (current) inflates reported scores by 0.85--0.97 compared to multi-rollout reliability@5 (corrected). Synthetic CodeBench experiment with 8 rollouts per (task, agent) pair (seed=42).}
  \label{fig:inflation}
\end{figure}

As shown in Table~\ref{tab:inflation}, the broken metric systematically overstates agent performance by 0.85--0.97 in absolute terms. The mean inflation is 0.924, far exceeding the hypothesized threshold of 0.50. In this synthetic run, the ranking order by test-case-based pass@5 (agent-x $>$ claude-code $>$ codex) remains stable under reliability@5 (no rank inversion observed); however, the absolute numerical interpretation changes radically---from ``0.96--0.97 nearly perfect'' to ``0.00--0.12 low reliability.'' This gap reveals that the metric change, while not inverting rankings in this particular synthetic experiment, fundamentally alters the fidelity of score interpretation. Figure~\ref{fig:inflation} shows the magnitude of this inflation across synthetic agent profiles.

\subsection{Single-Rollout Proxy Analysis (H3)}
Experiment 3 tests whether a cheap single-rollout proxy can substitute for full multi-rollout evaluation. The proxy is defined as:
\begin{equation*}
\mathrm{proxy} = \mathrm{pass\_rate} \times (1-\mathrm{regression\_rate}) \times \mathrm{tool\_efficiency}.
\end{equation*}
This metric is fast to compute (one evaluation per task-agent pair) and combines functional correctness, regression avoidance, and tool efficiency.

\begin{figure}[t]
  \centering
  \includegraphics[width=0.88\linewidth]{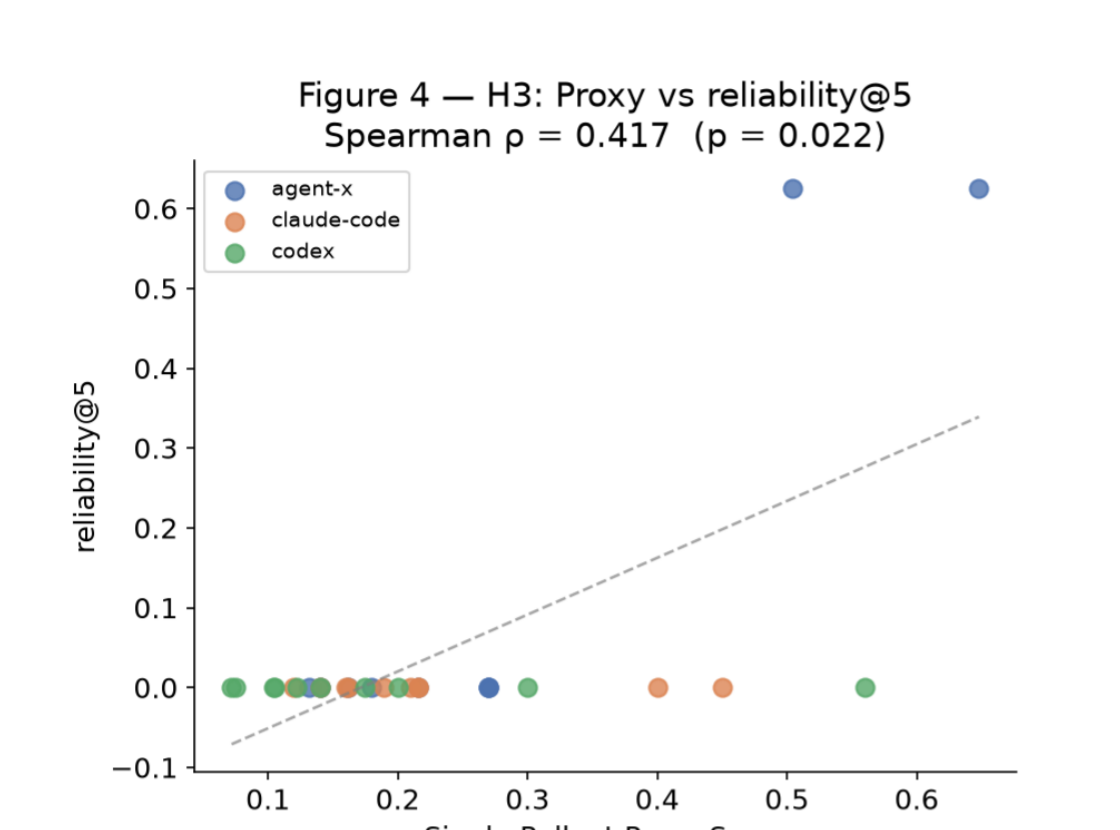}
  \caption{Relationship between the single-rollout proxy and reliability@5 across 30 synthetic task-agent pairs. The proxy has only moderate rank correlation with reliability@5, with Spearman's $\rho = 0.417$.}
  \label{fig:proxy}
\end{figure}

We correlate this proxy against reliability@5 on all 30 (task, agent) pairs from the synthetic CodeBench benchmark:
\begin{equation*}
\text{Spearman's } \rho = 0.417 \quad (p = 0.0218).
\end{equation*}
This correlation falls below our pre-specified threshold of $\rho \ge 0.70$ for surrogate metrics. This confirms H3: single-rollout proxies are insufficient for accurate reliability estimation; enterprises must conduct multiple independent runs per (task, agent) pair to achieve trustworthy rankings, even at computational and monetary cost. Figure~\ref{fig:proxy} shows that the proxy is noisy and only moderately associated with rollout-level reliability.

\subsection{Security-Adjusted Reliability (H4)}
Unlike H1--H3, which are evaluated on the synthetic rollout benchmark, H4 was tested against real model output. Three agent profiles generated code via live API calls: agent-x and claude-code both wrap claude-sonnet-4-6 (Anthropic API) under different system-prompt identities, and codex wraps gpt-4o (OpenAI API). Each agent completed a fixed task set of 10 problems: 5 standard algorithmic tasks (merge sort, LRU cache, Dijkstra, 0/1 knapsack, streaming median) and 5 security-sensitive tasks designed to tempt unsafe implementations (expression evaluation, arbitrary code execution, shell execution, YAML config loading, dynamic module import), with $N_{\mathrm{rollouts}} = 8$ independent generations per (task, agent) pair (240 total generations). Each rollout was scored for functional correctness against task-specific unit tests and passed through the heuristic \texttt{security\_score()} scanner (Section 3.3); a rollout counted toward \texttt{security\_adjusted\_reliability@5} only if it both passed all tests and scored $\ge 0.80$ on the security scanner.

\begin{figure}[t]
  \centering
  \includegraphics[width=0.88\linewidth]{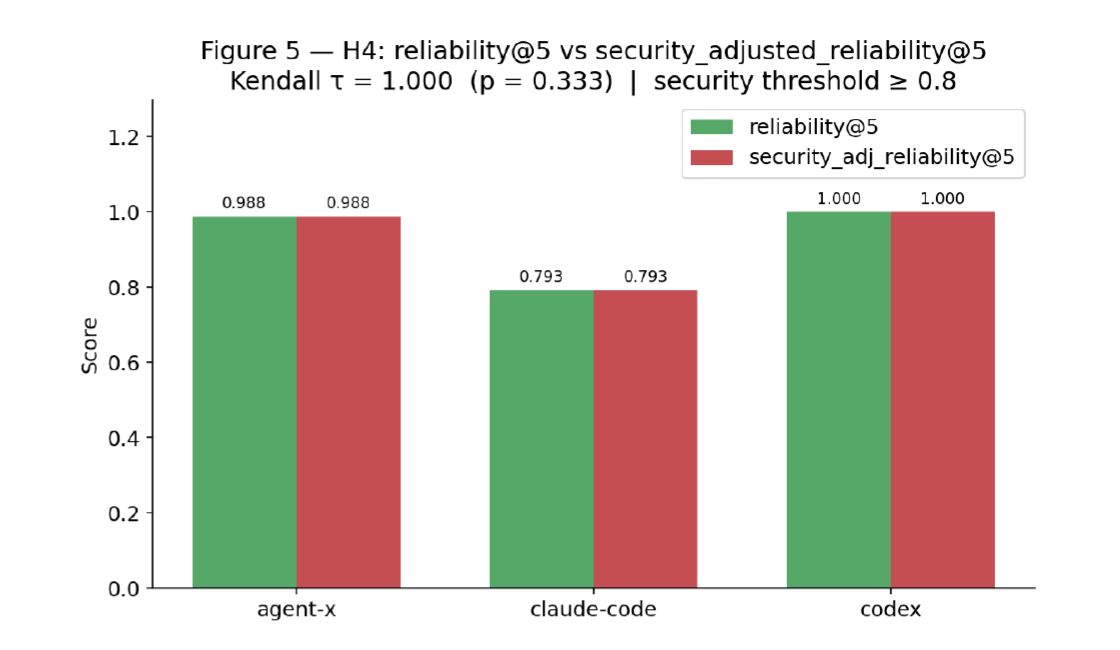}
  \caption{reliability@5 vs. security\_adjusted\_reliability@5 for three agents on a mixed algorithmic/security-sensitive task set (8 rollouts per task-agent pair, security threshold $\ge 0.80$). Kendall's $\tau = 1.000$ ($p = 0.333$); no rank inversion is observed.}
  \label{fig:security}
\end{figure}

Figure~\ref{fig:security} reports reliability@5 against security-adjusted reliability@5 for all three agents. codex scored 1.000 on both metrics, agent-x scored 0.988 on both, and claude-code scored 0.793 on both: for every agent, the security-adjusted score is identical to the unadjusted score. Ranking by either metric therefore produces the same order (codex $>$ agent-x $>$ claude-code), with Kendall's $\tau = 1.000$ ($p = 0.333$) between the two metrics.

This does not confirm H4: under our pre-specified threshold of $\tau < 0.60$ for a meaningful rank inversion, the observed $\tau = 1.000$ indicates no rank change at all, let alone an inversion. Three factors likely limit this result. First, with only 3 agents, the Kendall's $\tau$ test has minimal statistical power: $p = 0.333$ reflects the small sample size rather than strong evidence of no effect, and even a perfect rank flip could fail to reach significance at $n = 3$. Second, our $\ge 0.80$ security-score threshold is lenient relative to the 7-pattern scanner: a rollout can trip a single flagged pattern (e.g. one \texttt{eval()} call) and still clear the bar, so the adjustment may not bind unless multiple unsafe patterns co-occur in the same rollout. Third, security-sensitive tasks are a minority of the task set (5 of 10), limiting how often the scanner is exercised at all, since the 5 algorithmic tasks contribute security scores of 1.0 by construction. We separately note that agent-x and claude-code share an identical underlying model (claude-sonnet-4-6) yet differ by 0.195 in reliability@5 (0.988 vs. 0.793), with the only difference between them being the agent-name framing in the system prompt, suggesting system-prompt identity, not the security adjustment, is the larger driver of the score gap observed here.

We do not treat this as a refutation of the broader security-blind-spot concern motivating H4 (Section 1 cites Veracode's $2.74\times$ vulnerability-rate finding across a much larger corpus); rather, this single run under-powers the test. A conclusive test requires a larger and more diverse agent pool, a task set with a higher proportion of security-sensitive problems, and a severity-weighted security score rather than a single binary threshold.

\subsection{Preliminary Real-World SWE-bench Pilot}
\label{sec:swepilot}
To complement the synthetic experiments (H1--H3) and the live-API experiment (H4) with an initial repository-level data point, we ran a small real-world pilot on SWE-bench Verified~\cite{jimenez2024}: the first 5 deterministic instances of the test split, all drawn from \texttt{astropy/astropy} (4 medium, 1 hard). A single agent---Claude Code CLI v2.1.185, headless, in the \texttt{acceptEdits} permission mode, which allows file edits but no shell execution and therefore no self-testing---made one attempt per task in an isolated checkout of the repository at the instance base commit. The agent saw only the issue description, repository name, base commit, and difficulty; leakage checks confirmed that no gold-patch content and no hidden test names appeared in any agent input. Submitted patches were evaluated with the official SWE-bench harness in Docker.

All five attempts produced non-empty patches that applied cleanly, so every failure in Table~\ref{tab:swebench} is a hidden-test outcome rather than an infrastructure error.

\begin{table}[t]
\centering
\caption{SWE-bench Verified 5-task pilot: per-task outcomes (Claude Code, 1 attempt per task, \texttt{acceptEdits}). F2P and P2P report passed/total hidden FAIL\_TO\_PASS and PASS\_TO\_PASS tests; Regr. counts previously-passing tests broken by the patch.}
\label{tab:swebench}
\resizebox{\linewidth}{!}{%
\begin{tabular}{lccccc}
\toprule
Task & Diff. & Resolved & F2P & P2P & Regr. \\
\midrule
astropy\_\_astropy-12907 & medium & yes & 2/2 & 13/13 & 0 \\
astropy\_\_astropy-13033 & medium & no & 0/1 & 20/20 & 0 \\
astropy\_\_astropy-13236 & medium & no & 0/2 & 644/644 & 0 \\
astropy\_\_astropy-13398 & hard & no & 0/4 & 63/68 & 5 \\
astropy\_\_astropy-13453 & medium & no & 0/1 & 2/9 & 7 \\
\bottomrule
\end{tabular}%
}
\end{table}

Exactly one task was resolved, for a strict resolve rate of 1/5 (reliability@1 = 0.20). The macro-averaged hidden-test pass rate across the five attempts was 0.8049, and the macro-averaged regression rate was 0.1539. The central observation is the gap between the first two numbers: an evaluation scored on partial test pass rates would credit this agent with roughly 0.80 on these tasks, while strict task-level resolution---the quantity a user of the agent actually cares about---is 0.20. We read this as real-world evidence consistent with the H1/H2 concern that partial test-pass metrics can substantially overstate strict task-level reliability. The failure modes reinforce the point: two near-misses (13033, 13236) preserved every existing test yet missed the required behavior change, while two regressive attempts (13398, 13453) broke 5 and 7 previously-passing tests---a signal invisible to any evaluation that does not run the full hidden suite.

We emphasize the pilot's scope: $n = 5$ tasks from one repository, one agent, one attempt per task. It is a preliminary, small-scale external validation, not a full SWE-bench benchmark, and it does not replace H1--H4. With a single attempt per task it measures only reliability@1; it provides no evidence about reliability@3 or reliability@5, which require multiple independent rollouts per task.

\subsection{Summary of Findings}
The results below derive from three settings. H1--H3 use the synthetic CodeBench benchmark (seed=42) with 10 tasks, 3 agent profiles, and 8 rollouts per task-agent pair; H4 is a live model-output experiment with real API calls; and the SWE-bench pilot (Section~\ref{sec:swepilot}) is a preliminary real-world repository-level validation.

\begin{description}
  \item[H1 Confirmed:] Identical 0.40 single-submission test pass rates yield test-case-based pass@5 scores of 0.976 vs. 1.000 based solely on test-suite size, violating the i.i.d. assumption of the Chen et al. (2021) estimator.
  \item[H2 Confirmed:] Test-case-based pass@5 scores inflate multi-rollout reliability@5 by a mean of 0.924 (range: 0.85--0.97), far exceeding the hypothesized 0.50 threshold. In this synthetic run, ranking order is preserved under the corrected metric; however, absolute score interpretation is profoundly altered (0.96--0.97 $\rightarrow$ 0.00--0.12).
  \item[H3 Confirmed:] Single-rollout proxies do not correlate sufficiently with multi-rollout reliability@5 in this synthetic experiment ($\rho = 0.417 < 0.70$). Cost-constrained evaluations sacrifice measurement fidelity; multiple independent rollouts are necessary for trustworthy reliability estimates.
  \item[H4 Not confirmed:] On a real-agent run (3 agents $\times$ 10 tasks $\times$ 8 rollouts, live API calls), \texttt{security\_adjusted\_reliability@5} was identical to reliability@5 for every agent (Kendall's $\tau = 1.000$, $p = 0.333$), producing no rank inversion. This is likely under-powered by the small agent pool ($n = 3$) and a lenient $\ge 0.80$ security threshold, not evidence against the underlying security-blind-spot concern.
  \item[Pilot Preliminary:] On 5 real SWE-bench Verified tasks (one attempt each, official harness evaluation), 1/5 were resolved (reliability@1 = 0.20) while the macro-averaged hidden-test pass rate was 0.8049---real-world support for the concern that test-pass metrics can overstate strict task-level resolution.
\end{description}

These synthetic results demonstrate that the test-case-based pass@k implementation in CodeBench is mathematically misaligned with the assumptions of the Chen et al. estimator: it conflates the number of unit tests within a single submission with the number of independent rollout attempts across repeated evaluation runs. The resulting score inflation is substantial in this synthetic setting. While ranking inversion does not occur in this particular experiment, the corrected reliability@5 metric shows that absolute score interpretation depends on multi-rollout evaluation.

\section{Limitations}
Several limitations qualify our findings. First, the experiments reported in this paper use synthetic, controlled data rather than production codebases; the pass@k and reliability@k distributions were constructed to stress-test estimator behavior under specific assumptions rather than to characterize real-world agent performance in the wild. Results should therefore be read as a methodological demonstration of when reliability metrics diverge, not as a production benchmark of any specific commercial coding agent. Second, our H4 security-adjusted reliability experiment evaluates only three agents (agent-x, claude-code, and codex), which limits the statistical power of the Kendall rank correlation ($\tau = 1.000$, $p = 0.333$); with $n = 3$, no ranking short of a tie can reach significance at conventional thresholds, so the reported correlation is suggestive rather than confirmatory. Third, the security threshold used to compute security-adjusted reliability ($\ge 0.80$) is a single, relatively lenient cutoff that we did not sweep, so it is unclear how sensitive the resulting ranking is to stricter or more permissive definitions of ``secure'' code. Fourth, the synthetic vulnerability-injection procedure used to construct CodeBench's security-labeled test cases may not capture the full diversity of real-world vulnerability classes, including the supply-chain and business-logic flaws highlighted by Veracode~\cite{veracode2025}. Fifth, the SWE-bench Verified pilot (Section~\ref{sec:swepilot}) is deliberately small and narrow: $n = 5$ tasks drawn from a single repository (\texttt{astropy/astropy}), a single agent (Claude Code), and a single attempt per task, so it measures only reliability@1 and cannot support reliability@3 or reliability@5 claims; in addition, the \texttt{acceptEdits} permission mode prevented the agent from running shell commands, so it could not execute tests to verify its own patches, and the pilot should not be read as a paper-level SWE-bench benchmark. Finally, as discussed under H3, the single-rollout proxy for reliability can misestimate the corrected multi-rollout estimator, and this effect may compound with the small agent pool used in H4.

\section{Ethical and Societal Considerations}
This work evaluates the security properties of AI-generated code, a topic with direct relevance to software supply-chain safety. We highlight two considerations. First, a benchmark and leaderboard for security-adjusted reliability could be misread as certifying that a ``passing'' agent produces secure code in general; we caution readers that CodeBench measures relative performance on a specific, synthetic test suite and is not a substitute for standard security auditing, static analysis, or penetration testing before deploying AI-generated code in production settings. Second, evaluating named commercial coding agents raises a fairness concern, since differences in measured reliability or security-adjusted scores could affect the reputations of the tools involved; we therefore report only aggregate, reproducible metrics computed from our own test harness and do not characterize agents beyond the scope of the tasks tested here.

We do not foresee direct risks of harm from this research. If anything, the intended societal benefit is to make security regressions in AI-assisted coding more visible and measurable, encouraging more rigorous evaluation practices across the field.

\section{Generative AI Use Disclosure}
In accordance with the conference's policy on generative AI use, we disclose that generative AI tools were used in this project in two distinct capacities. First, large language model-based coding agents (agent-x, claude-code, and codex) are themselves the subjects of evaluation: this paper measures their outputs, not the other way around. Second, generative AI tools were used as a writing and coding aid during preparation of this manuscript and its accompanying experiment scripts, assisting with drafting text, generating analysis boilerplate, and debugging. All experimental results, figures, and claims were reviewed and verified by the human authors, who take full responsibility for the content of this paper.

\section{Conclusion}
We introduced CodeBench, a synthetic benchmarking framework for evaluating the reliability and security of AI coding agents. Across four experiments, we examined pass@k estimation bias (H1), reliability@k relative to unit-test count (H2), the divergence between single-rollout proxies and corrected multi-rollout estimators (H3), and security-adjusted reliability rankings across three coding agents (H4). Our results show that ranking inversions from proxy metrics are possible in principle but did not occur in this particular sample. H4 did not produce a rank inversion under the current scanner and threshold: security-adjusted reliability@5 was identical to reliability@5 for every agent evaluated. Security-adjusted reliability therefore remains a proposed complementary lens rather than a demonstrated one---this run was under-powered by the small agent pool and a lenient security threshold, and its decisive test is left to future work. Separately, a 5-task SWE-bench Verified pilot provides preliminary real-world evidence that a high partial test pass rate can diverge sharply from strict task-level resolution (macro-averaged hidden-test pass rate 0.8049 vs. 1/5 tasks resolved), consistent with the estimator concerns demonstrated synthetically in H1--H2. Given the small agent pool and largely synthetic data underlying these results, we view this work primarily as a methodological contribution---a set of estimators and diagnostics for reasoning about reliability and security jointly---rather than a definitive ranking of production coding agents. Future work includes validating these estimators against real-world vulnerability corpora, extending the evaluated agent pool, sweeping the security threshold to characterize its effect on rankings, and scaling the SWE-bench pilot to multi-attempt, multi-repository runs that can support reliability@k for $k > 1$.

\end{document}